# Epistemic Deference to AI

Benjamin Lange[1,2] [0000-0002-5809-8704]

[1] Ludwig-Maximilians-Universität München, Geschwister-Scholl Platz 1, 80539 München, Germany
[2] Munich Center for Machine Learning, Geschwister-Scholl Platz 1, 80539 München, Germany

**Abstract.** When should we defer to AI outputs over human expert judgment? Drawing on recent work in social epistemology, I motivate the idea that some AI systems qualify as Artificial Epistemic Authorities (AEAs) due to their demonstrated reliability and epistemic superiority. I then introduce AI Preemptionism, the view that AEA outputs should *replace* rather than supplement a user's independent epistemic reasons. I show that classic objections to preemptionism – such as uncritical deference, epistemic entrenchment, and unhinging epistemic bases – apply in *amplified* form to AEAs, given their opacity, self-reinforcing authority, and lack of epistemic failure markers. Against this, I develop a more promising alternative: a total evidence view of AI deference. According to this view, AEA outputs should function as *contributory* reasons rather than outright replacements for a user's independent epistemic considerations. This approach has three key advantages: *(i)* it mitigates expertise atrophy by keeping human users engaged, *(ii)* it provides an epistemic case for meaningful human oversight and control, and *(iii)* it explains the justified mistrust of AI when reliability conditions are unmet. While demanding in practice, this account offers a principled way to determine when AI deference is justified, particularly in high-stakes contexts requiring rigorous reliability.

## 1   Introduction

AI systems[1,2] increasingly outperform human experts.[3] For example, in medicine, AI systems can analyse images such as X-rays, MRIs, and CT scans more quickly and, in some cases, more accurately than doctors, detecting diseases like cancer at earlier stages and with greater precision.[4] Similarly, in finance, AI-driven trading algorithms execute high-frequency trades, optimize portfolio management, and forecast market trends with a level of speed and data-processing capability that surpasses human traders.[5]

---

[1]   I use the term 'AI' roughly for any machine or computer system capable of doing things that normally require intelligence and reflection – thinking, learning, or problem solving – when done by humans. See also Hasan et al. (2022, pp. 1-2).

[2]   I focus on *epistemic* AI systems (see Hauswald, section 1, forthcoming), which perform tasks previously reserved for human experts and epistemic authorities. Unlike AI designed for practical problem-solving, epistemic AI systems function primarily as sources of information or epistemic instruments, providing outputs that convey accessible worldly information (see Goldberg 2020, p. 2785).

[3]   According to Maslej et al's (2024) *AI Index Report 2024*, AI has surpassed human experts in tasks such as image classification, visual reasoning, language understanding, medical

As these AI systems continue to improve, a central question becomes: when should we *rationally* defer to AI's recommendations, particularly when they are more accurate than those of human expert authorities?

This question is not merely theoretical but has significant real-world consequences. Blind deference to AI could lead to over-reliance on systems that may harbour biases or hidden errors or that might compromise important epistemic virtues.[6] Conversely, irrational scepticism towards AI could lead to rejecting *better* judgments in favour of less reliable human reasoning. Striking the right balance requires a principled approach to *epistemic deference to AI*.[7]

In this paper, I develop a total evidence account of AI deference. According to this view, AI outputs should function as what philosophers call *contributory reasons* rather than preemptive replacements for human judgment. This approach situates the debate about when to rely on AI outputs in expert domains in the context of recent work in social epistemology, particularly discussions on deference to epistemic authority and expert testimony. I examine the implications of a total evidence account of AI deference, showing that it has three key advantages that align with existing concerns about the ethics and responsible use of AI: *(i)* it mitigates expertise atrophy by keeping human users epistemically engaged, *(ii)* it provides an *epistemic* rather than purely moral rationale for meaningful human oversight and control, and *(iii)* it explains justified epistemic mistrust of AI.

The article proceeds as follows. Section 2 formulates and motivates the case for Artificial Epistemic Authorities (AEAs) and AI preemptionism. I show that classic objections to Preemptionism – such as uncritical deference, epistemic entrenchment, and unhinging epistemic bases – apply in *amplified* form to AEAs, given their opacity, self-reinforcing authority, and lack of failure markers. Against these shortcomings, section

---

diagnosis, algorithmic efficiency, and materials discovery. However, it still struggles with complex reasoning, competition-level mathematics, and strategic planning.

[4] See Liu et al. (2019) for a comparison of AI and healthcare professionals in disease detection, Lebovitz et al (2021) for a critique of AI training and evaluation based on expert knowledge, and Han et al. (2024) for a review of AI in clinical practice. For a recent critical review of claims about AI outperforming human experts, see Drogt et al. (2024).

[5] For example, AI-driven algorithmic trading systems adjust to market information more quickly and may generate higher profits around news announcements due to their superior market timing ability and rapid execution (Bahoo et al., 2024; Frino et al., 2017).

[6] There are different questions about epistemic authority and deference. My focus here is the *deference question*: how we should rationally assign special epistemic weight to the views of an epistemic authority (once it is recognized). Other questions concern the *i)* nature and *ii)* identification of epistemic authority and *iii)* the transmission of epistemic goods (Jäger, forthcoming).

[7] The literature on epistemic deference to AI remains sparse, with Wolkenstein (2024) and Hauswald (2025; forthcoming) being among the few notable contributions that explicitly draw on social epistemology.

3 then proposes my total evidence account of AI deference, motivating its rationale and theoretic advantages. I show how this account can overcome some of the objections raised against AI preemptionism. Section 4 concludes by the implications of this account for current practice and usage of AI systems.

## 2 Artificial Epistemic Authorities and AI Preemptionism

We rely on others' expertise all the time. We trust doctors to diagnose and treat our illnesses, lawyers to interpret complex legal statutes, and auditors to ensure financial accuracy and compliance – often without understanding most or even any of the exact reasons and details. In doing so, we grant these experts a special kind of authority, recognising that their specialised knowledge places them in a superior epistemic position relative to us.[8]

What qualifies someone – or something – as an epistemic authority? According to *objectivist[9] belief-based views*, an epistemic authority must possess beliefs and the ability to communicate reasons for their judgments (e.g. Goldman, 2018). According to *objectivist functionalist views*, an epistemic authority must possess epistemic superiority such as being in a better position to provide reliable knowledge (e.g. Hauswald, 2024).

A promising definition of epistemic authority is provided by Jäger (2024):

> **Epistemic Authority[10]:** A is a (recognised) epistemic authority for person S in domain D at t and relative to epistemic goal G, if S truly believes A to be able, and prepared, to help S achieving S's epistemic goals in D and with respect to G, where this ability is due to A *de facto* being in a substantially advanced epistemic position in D, relative to S, at or during t, and with respect to the goal of acquiring G.

This view holds that A is an epistemic authority (EA) when a person S correctly believes that A has superior knowledge in a specific domain D at a given time t, relative to an epistemic goal G such as gaining knowledge or forming true beliefs. This means A must not only possess epistemic superiority but also be capable and willing to help S achieve their goal. For example, a patient might defer to a doctor for a diagnosis because they recognize the doctor's medical expertise.

---

[8] What is the relationship between expertise and epistemic authority? 'Expertise' and 'epistemic authority' need not be synonymous, as one may have authority without expertise and vice versa (Jäger, forthcoming). I here assume that epistemic authority should be conceived *relationally* and not in terms of an absolute threshold of epistemic knowledge. Experts, on the other hand, must cross an absolute epistemic threshold. They must be superior to most other people in the community with respect to D (Goldman, 2018).

[9] *Objectivist* accounts make a claim about the *de facto* epistemic conditions that an EA must fulfil. *Subjectivist* accounts maintain that an agent must *believe* or *perceive* a putative epistemic authority in an epistemically superior position.

[10] This is the most encompassing hybrid definition offered by Jäger (forthcoming), which requires actual epistemic superiority, recognition by a layperson, and a functional component that helps the layperson achieve their epistemic aims.

Given the above definition, we can ask whether *artificial* epistemic authorities (AEAs) could exist: AI systems that play a similar epistemic role to human epistemic authorities. There are good reasons to think that some AI systems might qualify as AEAs (Hauswald, 2025) – especially if we entertain the idea that epistemic authority concerns superior epistemic positioning and reliability rather than necessarily beliefs and communicative intent.[11]

First, many AI systems already provide helpful information, explanations, and reliable advice, functioning in ways that closely resemble traditional human EAs. From advanced medical diagnostic algorithms to financial risk analysis tools, AI is increasingly filling roles that require specialised epistemic expertise. Second, some AI systems hold superior epistemic positions in specific areas of applications. In such cases, AI systems do not merely replicate but arguably *surpass* human expertise, producing outputs that consistently reduce false positives and false negatives beyond human capability.[12] Third, the epistemic asymmetry between AI systems and human users in certain domains mirrors the traditional expert-layperson divide. For example, machine learning models trained on vast datasets can detect patterns and correlations that no human could reasonably discern, giving them an epistemic advantage akin to that of human experts over laypeople. Third, like human epistemic authorities, AI systems often exhibit epistemic opacity (Ross, 2024), so their reasoning processes, particularly in deep learning models, can be difficult or impossible for non-experts to fully grasp, even when they are highly reliable.[13]

If at least some AI systems qualify as AEAs, how ought we to respond to their outputs? In the case of human EAs, some have suggested that we should treat their judgments as *preempting* our own independent reasoning (Zagzebski, 2012; 2013; 2014; 2016).[14] Should we adopt a similar stance toward AEAs?

One answer is *AI Preemptionism*, which holds that when an AEA is a reliable truth-tracker, we should not merely add its outputs to our independent reasoning but replace our own reasons with its verdicts.

We can define this view more formally as

> **AI Preemptionism:** User U should treat the output O of a recognised artificial epistemic authority AEA in domain D at time t as a pre-emptive reason for believing proposition p, such that O replaces, rather than merely adds to, U's independent epistemic reasons for or against p.

---

[11] Some might be inclined to resist the idea of AEAs precisely because they can (currently) not meet criteria of beliefs and communicative intent. I am inclined to draw the opposite conclusion: because it seems intuitively compelling that AI-systems already serve as epistemic authorities, this provides a reason to reject these belief-based accounts.
[12] Maslej et al. (2024), chs. 2 and 5. See also footnote 3.
[13] For an insightful analysis about the tenuous relationship between transparency and expertise, see Nguyen C.T (2020).
[14] In addition to Zagyebski, Keren (2007, 2014a, 2014b) and Constantin and Grundmann (2018) have also defended forms of preemptionism for human EAs.

AI Preemptionism – like traditional Preemptionism – can be motivated through the so-called 'track-record' argument (Raz, 2009; Zagzebski, 2012). The basic idea is that if a putative epistemic authority reliably outperforms you, your best route to arrive at a correct view is to align with the epistemic authority who surpasses you in reliability and replace your own reasons with the authority's. Any 'half-measure' like adding or balancing your own evidence with the authority's output risks lowering overall accuracy in arriving at a true belief, which would be epistemically irresponsible.

Preemptionism has great appeal. It provides a clear and structured approach to epistemic deference which ensures individuals align their beliefs with the most reliable sources of knowledge. This is particularly compelling in cases where a layperson has no prior beliefs or reasons to rely upon or has conflicting beliefs or reasons that make an independent judgment difficult. In such situations, deferring entirely to an epistemic authority, rather than attempting to independently weigh competing evidence, maximises the chances of acquiring true beliefs while minimising the risk of getting it wrong.

The appeal of AI Preemptionism follows the same rationale but seems especially attractive because the AEA's potential for accuracy and reliability is so high. Unlike human experts, some AI-systems in narrow domains of application lack human-cognitive biases, fatigue, and other human cognitive inconsistencies, making its outputs arguably more precise and reproducible.[15] Moreover, AI systems can improve and self-refine their accuracy much more quickly than a human epistemic authority, whose expertise is only built up slowly over many years. AEAs also operates without social or economic pressures, potentially reducing conflicts of interest common among human epistemic authorities.

Though AI Preemptionism has an especially strong appeal, it inherits the classic objections to Preemptionism (Jäger, forthcoming). Indeed, I think that these objections are amplified in certain cases.

First, a standard objection to Preemptionism is that epistemic authorities are neither omniscient nor infallible and may therefore fail to consider all the reasons available to a non-expert.[16] This can lead to epistemic *losses* rather than gains. AEAs give this concern an additional problematic twist: the opacity of AI systems' reasoning, coupled with their empirical track record of superior performance in many domains, may risk systematically reinforcing *mistaken perceptions* of infallibility even when the AI lacks full epistemic access to all relevant considerations. Yet, due to their general statistical and predictive success, users may assume that the AI has already taken their own reasons into account, leading to a *stronger* epistemic pre-emption than would occur with a human authority. Thus, the concern is that AI Preemptionism strengthens the risk of uncritical deference, not merely because AI systems are epistemically superior in some respects but because they will always appear to be unquestionably so. This AI infallibility bias can leave users in a worse epistemic position than if they were deferring to a human expert, whose epistemic fallibility remains comparatively more detectable.

---

[15] Matters will depend on the specific AI-system in question. For example, insofar as current LLMs might qualify as AEAs in expert domains, suffer from other outputs bias compared to human EAs such as answer option variance in multiple-choice questions based on the position of the answer options (Pezeshkpour et al., 2023).

[16] See also Constantin and Grundmann (2020) and Grundmann (forthcoming).

A second objection to Preemptionism is that it can entrench individuals within irrational or epistemically corrupt communities, making it difficult for them to critically reassess misguided authorities whom they have been conditioned to trust (Lackey, 2018). This concern is even more acute regarding AEAs, as their epistemic authority can be systemic, automated, and self-reinforcing in ways that human authority is not. With human EAs, individuals often retain some capacity to challenge, interrogate, or disengage from a community's epistemic standards by identifying explicit biases, inconsistencies, or other irrational influences. However, AEAs can operate within algorithmically structured epistemic ecosystems that dynamically shape what users see, believe, and engage with, often in ways that are invisible to the users. This makes epistemic entrenchment more subtle. Also, unlike human EAs, whose reliability can be questioned based on incoherence, logical errors, or failures to adherence to established norms, standards, or methodologies within a given field of expertise, AEAs lack these clear epistemic failure markers. They do not exhibit hesitation, self-doubt, or accessible reasoning processes, which means that users embedded within AI-curated knowledge ecosystems may find it nearly impossible to recognise when their epistemic environment is corrupted. The danger is thus not merely that AI can reinforce irrationality but that it does so with a veneer of objectivity that is especially difficult to remove.

Third, Preemptionism has been criticised for requiring individuals to unhinge their own epistemic bases and replace them entirely with the authority's belief, which can lead to epistemic regress rather than progress (Jäger 2016; Dormandy 2018, Keren 2020). This concern is even more severe for AEAs. Unlike human EAs, current AI-systems do not provide reasons in the same way that humans do. Human EAs, even when opaque about specific justifications, operate within a shared epistemic framework that allows for some degree of reconstruction of their reasoning; by contrast, many AI systems, particularly those based on black box models, do not. However, if an AI system provides no explicit reasons for its belief but merely returns an output, then agents are expected to abandon their own epistemic bases without replacement. This makes the problem of epistemic regress even more acute. In medical AI, for instance, a doctor who pre-emptively defers to an AI's diagnostic decision might do so without understanding the features the AI weighed most heavily. If the doctor then ceases to consider their own medical reasoning in favour of the AI's opaque verdict, their epistemic base is not merely adjusted but entirely severed, leading to epistemic dependence without epistemic understanding.

These objections suggest that while AI Preemptionism might initially appear compelling – especially given AI's potential for superior reliability – it ultimately inherits and amplifies the classic problems associated with epistemic Preemptionism. AI systems, despite their computational advantages, are not epistemically infallible, and their authority can be dangerously opaque, systematically biased, and self-reinforcing in ways that human EA is not.

## 3     Total Evidence AI Deference Account

### 3.1    Total Evidence AI Deference

While AI Preemptionism offers a clear and structured approach to epistemic deference to AI, the concerns raised in the previous section highlight its significant limitations.

Total evidence views of epistemic authority may offer a promising alternative (Dougherty, 2014; Jäger, 2016; Lackey, 2018; Dormandy 2018).[17] Rather than advocating for complete pre-emption, total evidence views of epistemic authority hold that while authoritative testimony should carry significant epistemic weight, it should not completely replace an agent's independent reasons but rather be integrated as a *contributory reason* with an agent's own epistemic resources, allowing for aggregation rather than outright replacement of reasons.

Accordingly, we can define the AEA-relevant analogue of this view as follows:

> **Total Evidence AI Deference:** User U should treat the output O of a recognised artificial epistemic authority AEA in domain D at time t as a contributory reason for believing proposition p, such that O is integrated into, rather than replacing, U's independent epistemic reasons for or against p.

This definition contrasts with AI Preemptionism by allowing U to retain and weigh their own epistemic reasons rather than fully substituting them with AEA's output.[18]

There are different ways to spell out a total evidence view in detail. I here want to focus on a rendition that answers the concerns that emerged in the previous section. These can be formulated in terms of three desiderata.

First, a total evidence view of AI deference should acknowledge that while AI systems can serve as epistemic authorities, their reliability must be continuously assessed rather than assumed, to ensure that deference does not rest on an unwarranted assumption of infallibility. By preventing users from blindly trusting AI outputs simply because of their statistical success or predictive accuracy, this guards against uncritical deference. Second, the account must prevent epistemic entrenchment by ensuring that deference to AI remains an active, reflective process rather than a passive or self-reinforcing dependence. Given that AI systems can structure epistemic environments and reinforce specific viewpoints in ways that are difficult to challenge, users must retain the ability to critically assess AI-generated beliefs and recognise when deference may be leading them into an epistemically corrupt community. Third, the account should preserve independent epistemic bases by integrating AI-generated outputs with human reasoning to allow for epistemic progress.

We can formulate an account that meets these desiderate more precisely as:

---

[17]   See also Bokros (2021) for an interesting deference model in support of total evidence views based on formal accuracy-first epistemology.
[18]   For a similar argument in favour of using human second opinions to resolve disagreements between AI and human experts, see also Kempt and Nagel (2021).

**Total Evidence View of AI Deference**

**Critical Deference with Oversight:** Assume U is faced with the decision to belief p or non-p in domain D, then

1. U withholds of revisits deference to AEA if
    i. Domain Mismatch: p lies outside A's validated domain.
    ii. Reliability Undermining: Evidence suggests that A exhibits systematic bias or recurring errors.
    iii. Conflicting Authority: A comparably reliable human or AI EA disagrees.
    iv. Novel Evidence: U holds independent reasons that it is plausible that A did not consider.
2. Otherwise, U defers to AEA.

To illustrate, this accountant maintains that, by default, users should strongly defer to AEAs when it has a well-established epistemic advantage, but this deference remains *conditional* and *defeasible* rather than automatic. Unlike strict AI Preemptionism, which demands that users replace their own reasons with the authority's judgment, this model treats AI as an authoritative yet fallible epistemic agent whose outputs should be integrated into, rather than wholly substituted for, an agent's reasoning. The account thereby ensures that recognition of epistemic authority is based on demonstrated reliability, not on an uncritical assumption of infallibility. This means that while users should give significant epistemic weight to AI recommendations, they must also remain attuned to potential gaps in the AI's reasoning, limitations in its training data, or structural biases embedded in its design.

Moreover, this account prevents epistemic entrenchment by ensuring that AI deference remains open to revision. If users defer to AI uncritically, they risk being passively shaped by AI-driven knowledge ecosystems, making it difficult to recognise alternative perspectives, challenge embedded biases, or identify systematic epistemic distortions. To counter this, the model incorporates defeaters[19] and overrides that allow users to withdraw deference when clear signs of systematic bias, domain mismatch, or contradictory epistemic authority testimony arise.

Finally, this account also ensures that users do not become severed from their own epistemic bases. The account presented here avoids this by allowing for partial aggregation rather than full replacement. If a user identifies relevant, independent reasons that the AI might not have considered, those reasons can be reintroduced into the evaluation process, either by prompting the AI for an updated assessment or by weighing the user's own reasoning against the AI's conclusion. This ensures that epistemic progress is made rather than lost, maintaining a user's active role in belief formation rather than reducing them to a passive recipient of AI outputs.

---

[19] These can be more precisely understood as *rebutting defeaters*, which provide counterevidence, or *undercutting defeaters*, which question the reliability of the evidence itself. This contrasts with Constantin and Grundmann's (2018) *source-sensitive defeaters* which are specific to the credibility of the source of one's evidence.

## 3.2 Human Expertise Atrophy

One worry about deferring to AI is that human experts – doctors, lawyers, engineers – may gradually lose their domain-specific competences.[20] If a specialist consistently defers to an AI's outputs, they might become a passive conduit for algorithmic decisions rather than an active decision-maker, ultimately losing the skill to check or challenge the AI's recommendations. This risk of 'expertise atrophy' could hence render practitioners less capable of responding effectively if the AI performs poorly or faces unexpected scenarios outside its training scope.

The AI Deference Account helps mitigate this concern by demanding a *partial integration* of human reasoning and AI outputs, rather than total replacement. Since deference is *defeasible*, human experts must remain sufficiently active in the process to identify potential defeaters, domain mismatches, or updated evidence that the AI system may not have factored in. In practice, this encourages continuous skill development and preserves a meaningful epistemic role for humans, ensuring that domain knowledge and critical thinking do not wither away. Rather than undermining expertise, the AI Deference Account therefore suggests a dynamic in which AI complements rather than supplants human skill.

## 3.3 Meaningful Human Control

A second advantage of the AI Deference Account is that it provides an *epistemic* case for maintaining human control, independent of any moral or political values. While normative arguments for human oversight emphasise accountability, responsibility, and the preservation of autonomy (Amoros et al., 2020), the proposed accounts highlight the value of human input in identifying epistemic defeaters. Because the AI's recommendations are only *default-accepted* under conditions of demonstrated reliability, situations can and do arise where its authority is overridden, which necessitates a human arbiter to resolve conflicts, domain errors, or contradictory signals.

This requirement dovetails with existing accounts for responsible AI governance that advocate human oversight or 'meaningful human control' (see Santonio et al., 2018).[21] By explicitly incorporating checkpoints for re-evaluation, the AI Deference Account ensures that human users retain ultimate epistemic responsibility and are poised to step in whenever evidence emerges that the AI is mistaken or operating outside its domain. This not only protects against overreliance on automated systems but also aligns with regulatory and organisational guidelines promoting transparency and accountability in AI deployment.

---

[20] See Tai (2020) who discusses how automation leads to skill degradation and emphasizes the need for human oversight to prevent dependence on AI. Messeri and Crockett (2024) warn about "illusions of understanding", where human experts falsely believe they still possess expertise despite AI taking over complex tasks.

[21] On meaningful-human control (MHC), see Amoros et al. (2020), Verdiesen et al. (2021), and Mecacci et al. (2024). See Mosqueira-Rey et al. (2023) on human-in-the-loop approaches.

### 3.4   Lack of Epistemic Trust

A third benefit of the AI Deference Account is that it sheds light on why some users epistemically distrust certain AI systems.[22] According to this view, the rationality of trust in AI is grounded in the AI's capacity to meet the conditions for recognised epistemic authority (demonstrated reliability, domain alignment, transparency about its limitations, etc.). If these conditions remain unmet – for instance, due to opaque training data, a poor track record, or repeated biases – then it is epistemically *warranted* for users to withhold deference and maintain scepticism about the system's outputs.[23]

This account clarifies that user mistrust can be reasonable from a purely epistemic perspective, rather than constituting a knee-jerk rejection to technological innovation or status quo bias. If the AI cannot demonstrate that it holds a reliably superior vantage point in the relevant domain, or if it fails to accommodate defeaters that a competent human recognises, then deferring to it becomes irrational. The proposed account thereby validates and explains users' reluctance when conditions for justified deference are not met, providing a structured epistemic rationale for distinguishing between justified acceptance of AI outputs and rightful epistemic mistrust.

## 4   Implications for Current Use of AI

An implication of this account is that *few* current AI systems will easily satisfy the requirements needed to qualify as Artificial Epistemic Authorities (AEAs). At a minimum, systems must demonstrate sustained accuracy in their domain, provide a mechanism for ongoing reliability checks, and afford users the means to detect when defeater conditions arise. Meeting these benchmarks may demand substantial effort: designing interpretable architectures, collecting extensive domain-specific data that is continuously updated, and implementing transparent reporting of potential errors or biases. Some might consider these requirements unrealistic for large-scale or commercial AI deployments, especially under market pressures to release products quickly and cost-effectively. Nonetheless, where safety or high-stakes decision-making is crucial – as in medical diagnostics, finance, or aerospace – implementing these more stringent standards is both ethically *and epistemically* warranted.

Despite these hurdles, certain real-world AI use cases *do* already approximate the conditions for justified deference. Narrowly focused medical imaging AIs, for instance, can come close when they undergo rigorous clinical validation, operate within a well-defined domain, and offer clinicians a structure for second-guessing anomalous results. Similarly, some specialised AI systems in finance or engineering are subject to continuous audit and paired with expert human oversight, creating an environment in which defeaters can be systematically tracked.

---

[22]   Drawing on Wilholt's (2013) definition, epistemic trust refers to trust in AI strictly in virtue of its ability to provide reliable information or enhance human knowledge (Alvarado, 2023).
[23]   This aligns with Durán and Jongsma's (2021) argument that trust in AI does not necessarily require transparency but rather hinges on the reliability of its outputs.

In conclusion, I have examined whether and how we might justifiably defer to AI outputs rather than relying on our own or others' human judgment. I began by outlining AI Preemptionism, according to which an AI's recommendation *replaces* all of a user's existing reasons. Although Preemptionism appears compelling, classic objections to epistemic Preemptionism are amplified in AI contexts, given opaque algorithms, perceived infallibility, and the risk of reinforcing epistemically corrupt environments. To address these concerns, I proposed a total evidence view of AI Deference that treats AI outputs as *contributory* rather than *pre-emptive* reasons. By requiring ongoing reliability assessments, recognising defeaters, and retaining partial user agency, this approach avoids expertise atrophy, provides an epistemic rationale for human oversight, and explains users' justified distrust when AI systems fail to meet necessary epistemic conditions.

**Acknowledgments.** I thank audience members at the 2024 AISoLa conference as well as Frederic Gerdon, Sven Nyholm, and two anonymous referees and the editor from this journal for their feedback on earlier drafts of this paper. Special thanks to Johannes Reisinger for extensive comments.

**Disclosure of Interests.** The authors have no competing interests to declare that are relevant to the content of this article.

Deference to AI    13